\documentclass[twoside]{article}
\usepackage{amsfonts,amssymb,amsbsy,textcomp,marvosym,picins,amsmath,caption,threeparttable,amsthm,subfigure,float,lastpage,lscape}
\usepackage{eurosym,mathrsfs,fancyhdr,CJK,multicol,graphics,indentfirst,color,bm,upgreek,booktabs,graphicx,multirow,warpcol,epstopdf}
\usepackage[bookmarksnumbered=true,bookmarksopen=true,colorlinks=true,pdfborder=001,urlcolor=blue,linkcolor=blue,anchorcolor=blue,citecolor=blue]{hyperref}

\def\footnoterule{\kern 1mm \hrule width 10cm \kern 2mm}

\allowdisplaybreaks
\catcode`@=11
\def\title#1{\vspace{3mm}\begin{flushleft}\vglue-.1cm\Large\bf\boldmath\protect\baselineskip=18pt plus.2pt minus.1pt #1
\end{flushleft}\vspace{1mm} }
\def\author#1{\begin{flushleft}\normalsize #1\end{flushleft}\vspace*{-4pt} \vspace{3mm}}

\def\jz#1#2{{$^{\footnotesize\textcircled{\tiny #1}}$\let\thefootnote\relax\footnotetext{\!\!$^{\footnotesize\textcircled{\tiny #1}}$#2}}}
\catcode`@=11
\def\section{\@startsection{section}{1}{\z@}%
 {-3ex \@plus -.3ex \@minus -.2ex}%
 {2.2ex \@plus.2ex}%
{\normalfont\normalsize\protect\baselineskip=14.5pt plus.2pt minus.2pt\bfseries}}
\def\subsection{\@startsection{subsection}{2}{\z@}%
 {-3ex\@plus -.2ex \@minus -.2ex}%
 {2ex \@plus.2ex}%
{\normalfont\normalsize\protect\baselineskip=12.5pt plus.2pt minus.2pt\bfseries}}
\def\subsubsection{\@startsection{subsubsection}{3}{\z@}%
 {-2.2ex\@plus -.21ex \@minus -.2ex}%
 {1.4ex \@plus.2ex}
{\normalfont\normalsize\protect\baselineskip=12pt plus.2pt minus.2pt\sl}}

\begin{document}
\begin{CJK*}{GBK}{song}
\thispagestyle{empty}
\vspace*{-13mm}
\noindent {\small Data and system perspectives of sustainable artificial intelligence.%JOURNAL OF COMPUTER SCIENCE AND TECHNOLOGY
}
%===========================================================
\vspace*{2mm}

\title{Data and System Perspectives of Sustainable Artificial Intelligence}

% author names and affiliations
% use a multiple column layout for up to two different
% affiliations

    % Type all authors' full names above.
    % For a Chinese author, Chinse name is needed as well
    % Please declare the membership when any author is a member or fellow of CCF, ACM, IEEE

\author{Tao Xie$^1$, David Harel$^2$,  Dezhi Ran$^1$, Zhenwen Li$^1$, Maoliang Li$^1$, Zhi Yang$^1$, Leye Wang$^1$, Xiang Chen$^1$, Ying Zhang$^1$, Wentao Zhang$^1$, Meng Li$^1$, Chen Zhang$^3$, Linyi Li$^4$, Assaf Marron$^2$}
    % Type all authors' full names above.
    % For a Chinese author, Chinse name is needed as well
    % Please declare the membership when any author is a member or fellow of CCF, ACM, IEEE

%\address{
\noindent{}$^1$Peking University, Beijing, China. $^2$Weizmann Institute of Science, Rehovot, Israel. $^3$Shanghai Jiaotong University, Shanghai, China. $^4$Simon Fraser University, Canada.
%}

%\email{taoxie@pku.edu.cn,\{zhanglu,xiongyf,haod\}@sei.pku.edu.cn; xxiao2@ncsu.edu}
%\received{month day, year}
%\revised{month day, year}%Revised date

\let\thefootnote\relax\footnotetext{{}\\[-4mm]\indent\ Regular Paper}

\noindent {\small\bf Abstract} \quad  {\small %\textcolor{blue}
Sustainable AI is a subfield of AI for concerning developing and using AI systems in ways of aiming to  reduce environmental impact and achieve sustainability. Sustainable AI is increasingly important given that training of and inference with AI models such as large langrage models are consuming a large amount of computing power. In this article, we discuss current issues, opportunities and example solutions for addressing these issues, and future challenges to tackle, from the data and system perspectives,  related to data acquisition, data processing, and AI model training and inference.}

\vspace*{3mm}


\vspace*{4mm}

\end{CJK*}
\baselineskip=18pt plus.2pt minus.2pt
\parskip=0pt plus.2pt minus0.2pt
\begin{multicols}{2}

\section{Introduction}

Sustainable AI is a subfield of AI for concerning developing and using AI systems in ways of aiming to  reduce environmental impact and achieve sustainability. Sustainable AI is increasingly important given that training of and inference with AI models such as large langrage models are consuming a large amount of computing power. In this article, we discuss current issues, opportunities and example solutions for addressing these issues, and future challenges to tackle, from the data and system perspectives,  related to data acquisition, data processing, and AI model training and inference.

\section{Data Acquisition}
% to be written by Zhang Ying and Wang Leye

\subsection{Current Issues for Sustainable AI}

As artificial intelligence (AI) technologies continue to proliferate across industries, data acquisition plays a crucial role in the development and training of AI models. However, current data acquisition methods face several challenges, particularly in terms of environmental impact, privacy concerns, data quality, and compliance, all of which directly influence the sustainability of AI systems.
\subsubsection{Environmental Cost and Energy Consumption}
Data acquisition and AI model training, especially for large-scale machine learning models like deep learning, require significant computational resources and energy. Studies have shown that training a medium-sized generative AI model can consume energy equivalent to the lifetime carbon emissions of five average American cars. This issue becomes particularly concerning when large amounts of data are required, significantly increasing the environmental footprint of AI. The need to address energy efficiency in AI systems is becoming increasingly urgent, as training AI models at scale can cause severe environmental impacts if not managed properly \cite{strubell2019}.

\subsubsection{Data Opacity and Quality Issues}
While many AI systems rely on publicly available datasets, the lack of transparency and the quality of data remain significant challenges. Public datasets often suffer from issues like bias, mislabeling, and lack of representativeness, undermining the accuracy and fairness of trained models. Moreover, the lack of transparency in data sources makes it difficult to track data provenance, which can lead to the propagation of biased or misleading outcomes in AI applications \cite{buolamwini2018}. Improving data quality and transparency is a critical aspect of ensuring sustainable AI development.

\subsubsection{Underutilization of Private Data}
Currently, much of the valuable private data (such as user behavior data and personalized information) remains underutilized. Compared to public data, private data often provides higher-quality and more personalized insights. However, the use of private data is heavily constrained by privacy protection laws and user consent regulations \cite{shokri2017}. The challenge lies in maximizing the value derived from private data while ensuring compliance with privacy regulations and maintaining user trust. This represents a key hurdle in advancing sustainable AI.

\subsubsection{Underexploited Non-Textual Data}
AI models today primarily rely on textual data for training, but significant opportunities remain in the underutilized non-textual data, such as acoustic data, spectral data, and sensor data. These forms of non-textual data hold immense potential in fields like environmental monitoring, healthcare diagnostics, and precision agriculture \cite{sun2017}. However, the collection and processing of non-textual data are often costly and technically challenging. Despite these hurdles, the potential of such data to contribute to sustainable AI development remains largely untapped.

\subsubsection{Data Privacy Concerns}
Data privacy has always been a sensitive issue in AI development, particularly when dealing with sensitive personal information such as biometric data or location information. Whether it is public data or private data, ensuring compliance with privacy laws (such as GDPR) and protecting users' rights is a critical consideration. Even anonymized or de-identified data can pose privacy risks due to the possibility of re-identification \cite{narayanan2008}. This makes the balance between data utilization and privacy protection a major challenge for sustainable AI.

\subsection{Opportunities and Example Solutions for Addressing These Issues}
Despite the aforementioned challenges, there are numerous opportunities to improve data acquisition processes in AI systems, ensuring their sustainability and compliance with environmental, ethical, and regulatory standards.

\subsubsection{Cost-effective Data Acquisition (By Wentao)} 
Crowdsourcing has become a widely adopted method for cost-effective data acquisition, particularly for tasks requiring large-scale data annotation. However, crowdsourcing introduces challenges related to the quality of annotations, as the speed-accuracy trade-off described by Fitt's law often results in noisy or biased labels, especially for ambiguous or borderline cases. To address these issues, researchers have developed techniques such as collaborative labeling, where redundant annotations from multiple annotators are aggregated using consensus strategies. Additionally, methods like truth inference~\cite{zheng2017truth} and iterative information rejection~\cite{xie2012iterative} have been proposed to improve the reliability of crowdsourced data. These approaches help mitigate the risks of low-quality annotations, making crowdsourcing a scalable and sustainable solution for data acquisition in AI development.

Active learning (AL) provides a strategic approach to data acquisition by prioritizing the most informative data samples for annotation, thereby reducing the overall cost and effort required for training AI models. The AL process begins with a small set of labeled data, which is used to train an initial model. Based on the model's predictions, priority scores are assigned to unlabeled data samples, and the most valuable samples are selected for annotation using various query heuristics. This iterative process continues until the model achieves satisfactory performance. By focusing on the most informative and representative data, active learning not only reduces the cost of data acquisition but also enhances the efficiency and sustainability of AI model training.

\textbf{Example Solution:} Platforms such as Amazon Mechanical Turk (MTurk) and LabelBox enable organizations to distribute labeling tasks to a global workforce, significantly reducing the time and cost associated with manual data annotation. Similarly, companies like Google have integrated active learning into their data pipelines to optimize the annotation process for large datasets, further demonstrating its potential for cost-effective and sustainable data acquisition.

\subsubsection{Improving Energy Efficiency in AI Systems}
Reducing the energy consumption of AI training and data processing is paramount for making AI more sustainable. Several promising strategies can be employed to achieve this, including the use of energy-efficient hardware, optimization of algorithms to reduce computational complexity, and the adoption of renewable energy sources to power data centers. Additionally, distributed and edge computing presents an opportunity to offload computational tasks to local devices, reducing the need for centralized data processing and lowering energy usage.

\textbf{Example Solution:} Google, for instance, has significantly reduced the energy consumption of its data centers by using AI algorithms to optimize cooling systems, lowering both energy consumption and carbon emissions. Such initiatives offer valuable insights for other organizations aiming to reduce the environmental impact of their AI systems \cite{gubbi2013}.

\subsubsection{Enhancing Data Transparency and Quality}
To mitigate the issues surrounding data quality and transparency, the implementation of more robust data management frameworks is critical. This includes the use of automated data cleaning and validation techniques, as well as the adoption of standardized protocols for data collection and processing. Blockchain technology, in particular, offers a promising solution for improving data traceability and ensuring transparency in data provenance, enabling stakeholders to verify the source and integrity of datasets \cite{wood2014}.

Another powerful approach is data synthesis, which involves generating synthetic data that mimics real-world data without compromising privacy. Data synthesis can be especially useful in situations where access to sensitive or limited data is an issue. For example, in healthcare or finance, synthetic data can be used to train AI models while protecting personal information. By generating realistic datasets that preserve the statistical properties of real data, data synthesis can enrich training datasets and reduce the risks associated with data privacy \cite{franklin2020}. This approach can also address issues of data imbalance, where certain classes or outcomes are underrepresented in the dataset, improving the robustness and fairness of AI models.

\textbf{Example Solution:} Companies such as Waymo and NVIDIA leverage data synthesis for training their self-driving car models. Using simulated environments, these companies generate synthetic data representing various driving scenarios, weather conditions, and traffic situations that may be rare or difficult to capture in real-world datasets. By using computer-generated simulations, they can extensively test the robustness and safety of autonomous vehicle algorithms \cite{waymo2019}. Waymo, for example, has used synthetic data to improve object detection and reaction times of their self-driving systems, reducing the need for extensive real-world testing, which can be costly and time-consuming.

\subsubsection{Leveraging Private Data in a Privacy-Preserving Manner}
The potential of private data remains largely untapped due to concerns over privacy and regulatory compliance. New technologies such as federated learning, differential privacy, and homomorphic encryption offer innovative ways to utilize private data without compromising privacy \cite{mcmahan2017, dwork2008}. These techniques enable organizations to train AI models on sensitive data while ensuring that the data never leaves its original location or is exposed to unauthorized parties.

A promising framework for enabling secure and compliant data sharing is the International Data Spaces (IDS). The IDS initiative creates a secure, trusted data-sharing ecosystem where organizations can share and collaborate on data while maintaining control over their sensitive information. In this model, data remains within the respective organizational boundaries and is only accessible to authorized parties through secure, transparent mechanisms \cite{bohnenberger2021}.

The IDS relies on a set of guidelines and technical standards that facilitate the creation of data spaces-secure environments where data can be shared under predefined conditions. These data spaces ensure that data providers retain control over their data, granting access only to those who comply with the agreed-upon conditions. By incorporating these principles, the IDS aims to foster a secure, privacy-respecting environment for data sharing and collaboration, essential for enabling the use of private data in AI systems.

\textbf{Example Solution:} The HealthDataSpace project, part of the European Union's initiative, is a prime example of leveraging private data in a privacy-preserving manner using IDS principles. Through this project, hospitals and medical organizations can share anonymized patient data across borders for AI training without compromising privacy. By using the IDS framework, sensitive health data is kept within each organization's infrastructure, and only aggregated, anonymized, or pseudonymized data is shared under strict, consent-based agreements \cite{healthdataspace2021}.

\subsubsection{Utilizing Non-Textual Data}
Non-textual data, such as sound, light, and sensor data, represents an underexploited resource that holds vast potential for enhancing AI models. Advances in sensor technologies, coupled with AI's growing ability to process multimodal data, provide an opportunity to harness these data types for a wide range of applications. By integrating non-textual data with existing textual datasets, AI models can become more accurate, efficient, and adaptable.

\textbf{Example Solution:} In environmental monitoring, AI-powered sensors are being used to track deforestation, monitor wildlife, and measure air quality. These sensors collect valuable non-textual data that can be integrated with AI models to provide real-time insights into environmental changes and contribute to more sustainable practices \cite{xu2020}.

\subsubsection{Ensuring Data Privacy Protection}
As data privacy continues to be a major concern, privacy-preserving AI techniques will play a critical role in enabling the responsible use of data. Techniques like homomorphic encryption and secure multi-party computation allow AI systems to process encrypted data, ensuring that sensitive information remains private while still contributing to model training \cite{brakerski2014}.

\textbf{Example Solution:} By employing homomorphic encryption, AI systems can analyze data without decrypting it, thus ensuring that sensitive information remains confidential. Such advancements in privacy-preserving technologies will enable the ethical use of data, fostering trust and compliance with privacy regulations \cite{chen2020}.

\subsection{Future Challenges to Tackle}
Despite significant progress, several challenges remain that need to be addressed in the pursuit of truly sustainable AI. These challenges involve not only technological advancements but also regulatory, ethical, and organizational changes.

\subsubsection{Legal and Ethical Issues in Private Data Utilization}
As privacy laws continue to evolve, finding a balance between utilizing private data and ensuring compliance with regulations will remain a key challenge. International differences in data protection laws and the increasing complexity of privacy regulations will necessitate greater collaboration between governments, industry leaders, and AI developers to create standardized frameworks that protect user privacy while enabling data-driven innovation \cite{regan2015}.

\subsubsection{Standardization of Non-Textual Data}
The collection, processing, and integration of non-textual data are hindered by the lack of standardized formats and technical infrastructure. The development of universal standards for non-textual data, coupled with the creation of interoperable platforms for data sharing, will be crucial to unlocking the full potential of these data types in AI applications \cite{herzog2021}. Without common standards, it will be difficult to integrate diverse data sources effectively and create robust models that utilize multimodal inputs, limiting the overall scalability and performance of AI systems.

\subsubsection{Sustainable Scalability of AI Systems}
As AI systems become increasingly sophisticated and require larger datasets for training, ensuring their scalability in an environmentally sustainable manner will become a major challenge. The environmental impact of training ever-larger models with massive datasets will require continued innovation in algorithmic efficiency, energy-efficient hardware, and green data center infrastructure \cite{stoica2014}. Sustainable scaling of AI will be crucial in preventing AI's environmental footprint from expanding uncontrollably. Thus, finding ways to optimize training processes and improve energy efficiency at scale will be essential for achieving long-term sustainability in AI research and development.

\section{Data Processing}
% to be written by Zhang Wentao and Wang Leye
\subsection{Current Issues for Sustainable AI} 
Data processing is a critical component of sustainable AI development, as it directly impacts the quality, efficiency, and environmental footprint of AI systems. However, several challenges persist in current data processing practices, particularly in the context of improving data quality and efficiency.

\subsubsection{Data Cleaning Challenges}
Real-world datasets often contain noise, errors, and missing values due to the labor-intensive nature of data collection. These issues can arise in both features and labels, leading to poor model performance and increased computational costs. For example, noisy labels can introduce ambiguity during training, while missing values can reduce the representativeness of the dataset. Traditional data cleaning methods, such as rule-based filtering, are often insufficient to handle the complexity and scale of modern datasets~\cite{chu2016data}. Moreover, the lack of automated and scalable solutions for data cleaning exacerbates the problem, making it a significant bottleneck in sustainable AI development.

\subsubsection{Feature Engineering Challenges}
Feature engineering is essential for transforming raw data into meaningful representations that improve model performance. However, this process is often time-consuming and requires domain expertise, which can be a barrier for organizations with limited resources~\cite{kumar2017data}. In the era of deep learning, feature engineering is sometimes overlooked, as models are expected to automatically learn relevant features. However, this approach is less effective for structured data, where domain-specific features are crucial for achieving high accuracy. Additionally, irrelevant or noisy features can increase computational overhead and reduce model efficiency, further complicating the development of sustainable AI systems.

\subsubsection{Imbalanced Data Challenges}
Imbalanced data distributions are common in real-world applications, where one class significantly outnumbers the others. This imbalance can lead to biased models that prioritize the majority class, resulting in poor performance on minority classes~\cite{he2009learning}. For example, in defect detection systems, rare defects may be overlooked due to their low representation in the training data. Traditional methods for addressing imbalanced data, such as oversampling or undersampling, often fail to generalize well or introduce additional noise into the dataset. These challenges highlight the need for more robust and sustainable solutions to handle imbalanced data effectively.

\subsubsection{Data Selection Challenges}
While data selection aims to improve data efficiency by identifying valuable subsets from large datasets, it faces significant challenges. One major issue is the difficulty of defining criteria for selecting the most informative samples, especially in high-dimensional data spaces. Additionally, data selection methods often struggle to balance the trade-off between data reduction and model performance~\cite{kim2022defense}. For example, removing redundant or less informative samples may inadvertently exclude critical data points that are essential for capturing rare but important patterns. This can lead to biased or underperforming models, particularly in applications where data diversity is crucial, such as healthcare or autonomous driving.

\subsubsection{Data Augmentation Challenges}
Data augmentation is widely used to enhance dataset size and diversity, but it is not without its challenges. One key issue is the risk of introducing unrealistic or misleading data points during the augmentation process. For instance, in computer vision, overly aggressive transformations (e.g., extreme rotations or distortions) may generate images that do not reflect real-world scenarios, leading to poor model generalization. Furthermore, data augmentation techniques often require domain-specific knowledge to ensure that the augmented data retains the statistical properties of the original dataset~\cite{feng2021survey}. This can be particularly challenging in fields like natural language processing (NLP), where semantic consistency must be preserved during augmentation.

\subsection{Opportunities and Example Solutions for Addressing These Issues} 
Despite the challenges, there are numerous opportunities to enhance data processing techniques, making them more efficient, scalable, and environmentally friendly.

\subsubsection{Automated Data Cleaning}
Advancements in machine learning have enabled the development of automated data cleaning tools that can detect and resolve errors in large datasets~\cite{krishnan2019alphaclean}. For example, ML-based anomaly detection algorithms can identify noisy samples~\cite{le2022log}, while imputation techniques can handle missing values more effectively. These tools reduce the manual effort required for data cleaning and improve the overall quality of datasets, leading to more accurate and efficient AI models.

\textbf{Example Solution:}
OpenRefine~\cite{verborgh2013using}, an open-source tool, uses machine learning algorithms to clean and transform messy datasets. It can automatically detect inconsistencies, merge duplicate entries, and fill in missing values, making it a valuable resource for sustainable AI development.

\subsubsection{Automated Feature Engineering}
Automated feature engineering (AFE) tools leverage machine learning to generate relevant features from raw data, reducing the need for manual intervention. These tools can identify patterns and relationships in the data that may not be apparent to human engineers, leading to more robust and generalizable models. AFE is particularly useful for structured data, where domain-specific features are critical for model performance.

\textbf{Example Solution:}
Featuretools~\cite{kanter2015deep} is a popular AFE library that uses deep feature synthesis to automatically create features from relational datasets. By automating the feature engineering process, Featuretools reduces the time and expertise required to build high-quality models, contributing to the sustainability of AI systems.

\subsubsection{Advanced Techniques for Imbalanced Data}
To address the challenges of imbalanced data, researchers have developed advanced techniques such as synthetic data generation and cost-sensitive learning. Synthetic data generation methods, such as SMOTE (Synthetic Minority Over-sampling Technique)~\cite{chawla2002smote}, create realistic samples for minority classes, balancing the dataset and improving model performance. Cost-sensitive learning~\cite{zhou2005training}, on the other hand, assigns higher penalties to misclassifications of minority classes, encouraging the model to focus on underrepresented data.

\textbf{Example Solution:}
In healthcare, synthetic data generation has been used to address the imbalance in medical datasets. For instance, researchers have used SMOTE to generate synthetic samples of rare diseases, enabling AI models to achieve higher accuracy in diagnosing these conditions. This approach not only improves model performance but also reduces the need for extensive data collection, making it a sustainable solution for imbalanced data challenges.

\subsubsection{Advanced Data Selection Techniques}
Recent advancements in machine learning have led to the development of more sophisticated data selection methods. For example, active learning-based approaches can iteratively select the most informative samples for annotation, reducing the need for large labeled datasets~\cite{sener2017active}. Similarly, clustering-based methods can identify representative subsets of data that capture the diversity of the entire dataset. These techniques not only improve data efficiency but also reduce the computational resources required for training, contributing to the sustainability of AI systems.

\textbf{Example Solution:}
Google has implemented data selection techniques in its data pipelines to optimize the training of large-scale machine learning models. By identifying and removing redundant data points, Google has significantly reduced the computational cost of model training while maintaining high accuracy. This approach has been particularly effective in applications like image recognition and natural language processing, where data redundancy is common.

\subsubsection{Intelligent Data Augmentation}
To address the challenges of data augmentation, researchers have developed intelligent augmentation techniques that leverage machine learning to generate realistic and diverse data points. For example, AutoAugment~\cite{cubuk2019autoaugment} uses reinforcement learning to automatically discover optimal augmentation policies for specific datasets and tasks. This approach ensures that augmented data retains the statistical properties of the original dataset while maximizing diversity. Additionally, domain-specific augmentation techniques, such as synonym replacement in NLP or geometric transformations in computer vision, can enhance data quality without introducing unrealistic artifacts.

\textbf{Example Solution:}
In healthcare, researchers have used intelligent data augmentation to generate synthetic medical images for training AI models. By applying realistic transformations to existing images, they have created diverse datasets that improve model performance in tasks like tumor detection and disease classification. This approach reduces the need for extensive data collection and annotation, making it a sustainable solution for data augmentation.

\subsection{Future Challenges to Tackle} 
While significant progress has been made in improving data processing techniques for sustainable AI, several challenges remain that must be addressed to ensure the long-term viability and scalability of these solutions. 

\subsubsection{Ensuring Data Quality at Scale}
As datasets continue to grow in size and complexity, ensuring data quality becomes increasingly challenging. Noise, errors, and biases can propagate through large datasets, leading to poor model performance and unreliable predictions. Future research must focus on developing scalable and automated data cleaning techniques that can handle the volume and variety of modern datasets without compromising accuracy. Additionally, there is a need for robust frameworks to evaluate and monitor data quality in real-time, particularly in dynamic environments where data is continuously updated.

\subsubsection{Bridging the Data Quantity Gap}
Many application domains, such as healthcare and environmental monitoring, suffer from limited data availability due to high collection costs or privacy concerns. Synthetic data generation and data augmentation techniques offer promising solutions, but they must be further refined to ensure that generated data retains the statistical properties and diversity of real-world data. Future work should also explore the ethical implications of synthetic data, particularly in sensitive domains where data misuse could have serious consequences.

\subsubsection{Addressing Bias and Fairness}
Bias in datasets can lead to unfair or discriminatory outcomes, particularly in applications like hiring, lending, and law enforcement. While data cleaning and augmentation techniques can help mitigate bias, they are often insufficient to address systemic issues in data collection and labeling. Future work must focus on developing frameworks for identifying, quantifying, and mitigating bias in datasets, as well as ensuring that AI systems are transparent and accountable in their decision-making processes.

\subsubsection{Scalability in Resource-Constrained Environments}
Many data processing techniques, such as active learning and advanced data augmentation, are computationally intensive and may not be feasible in resource-constrained environments. Future work must focus on developing scalable and efficient algorithms that can be deployed in settings with limited computational resources, such as edge devices or low-income regions. This includes exploring distributed computing frameworks and techniques for reducing the memory and communication overhead of data processing tasks.

\subsubsection{Developing User-Friendly and Accessible Tools}
The complexity of data processing techniques often limits their adoption by non-experts, particularly in small organizations or resource-constrained settings. Future work must focus on developing user-friendly tools and platforms that simplify the implementation of data-centric approaches, making them accessible to a broader audience. This includes creating intuitive interfaces, providing comprehensive documentation, and offering pre-built solutions for common data processing tasks.

\section{AI Model Training and Inference}
% to be written by Yang Zhi, Zhang Chen, and Wang Chenxi

\subsection{Current Issues}
The sustainability of AI hardware architectures and system software stacks faces numerous challenges, including the mismatch between computational power and the needs of large-scale AI models, memory bandwidth and latency bottlenecks, hardware architecture heterogeneity, and insufficient development toolchains. Although current accelerators have made significant progress in terms of performance, they still face limitations, especially when it comes to large-scale training and inference. Addressing issues such as improving energy efficiency, reducing latency, enhancing cross-platform collaboration, and optimizing toolchains will be key to the future development of AI hardware architectures.

\subsubsection{Performance Bottlenecks in AI Model Training and Inference}
\textbf{Computational Power.} Current AI models, particularly deep learning models, require increasing computational power year by year due to the growing size of datasets and the increasing complexity of the models. For example, OpenAI's GPT-3 model, which has 175 billion parameters, requires nearly hundreds of Petaflops of computational power during training. Existing hardware architectures, such as NVIDIA's A100 GPU and Google's TPU v4, provide significant acceleration for large-scale computation tasks, but their computational power is still insufficient to meet the demands of increasingly large models. For instance, the NVIDIA A100 GPU has a peak floating-point performance of 312 teraflops (TFLOPS), yet when dealing with very deep neural networks, the computational load often exceeds the capability of a single accelerator, leading to bottlenecks and inefficient hardware resource utilization. Particularly in large-scale neural network training, the computational power of a single accelerator is inadequate to maintain efficient data processing speeds, resulting in substantial idle hardware resources and reduced training efficiency.

\textbf{Memory Bandwidth and Latency.} The processing of large datasets in AI training and inference depends heavily on efficient memory access and data transfer. However, memory bandwidth and latency have become critical bottlenecks that limit performance improvements. For example, the NVIDIA A100 GPU has a memory bandwidth of 1555GB/s, but when faced with large neural networks, this is still insufficient to support the frequent access to numerous parameters and data during training. In BERT model training, for instance, while the model itself has a smaller parameter size (110 million parameters), the large amount of intermediate data that needs to be read and written to memory causes memory bandwidth to become a performance bottleneck. Moreover, memory latency significantly impacts the efficiency of large-scale AI model inference. For deep neural networks with billions of parameters, frequent memory accesses cause high latency as the computing cores wait for data, leading to reduced overall inference performance.

\textbf{Energy Efficiency Issues.} As AI models continue to scale up, energy efficiency becomes an increasingly critical issue. Although current AI accelerators have made significant advancements in computational performance, improvements in energy efficiency have been insufficient. For example, the power consumption of the NVIDIA A100 GPU is around 400W, and despite this power consumption, it still cannot meet the needs of large-scale AI training. According to research by DeepMind, the energy consumption required to train the GPT-3 model is approximately 1280 MWh, equivalent to the electricity usage of a typical household over ten years. This indicates that, while computational power continues to increase, energy efficiency remains a significant barrier to achieving sustainable development in AI hardware, particularly when deploying AI tasks in large-scale cloud computing centers where the balance between energy efficiency and cost is especially prominent.

\subsubsection{Lack of Heterogeneity and Adaptability in Hardware Architectures}
\textbf{Challenges of Hardware Heterogeneity.} The diverse demands of AI training and inference have led to the heterogeneity of hardware architectures, but the coordination between different hardware architectures remains a pressing issue. Existing hardware architectures, such as CPUs, GPUs, FPGAs, and TPUs, each have distinct computational characteristics and advantages, but their integration and efficient collaboration face many challenges. For example, while GPUs excel at large-scale parallel computing, CPUs or FPGAs may be more efficient for certain AI tasks (e.g., processing small data streams). Traditional AI accelerators such as TPUs are designed specifically for deep learning training and inference, but they tend to perform poorly in more general-purpose computing tasks. Differences in resource scheduling, data sharing, and synchronization mechanisms between various hardware platforms make cross-platform cooperation complex, thereby impacting overall system computational efficiency and resource utilization.

\textbf{Architectural Adaptability.} Different AI application scenarios (e.g., image recognition, natural language processing, reinforcement learning) impose significantly different requirements on hardware architectures, which limits the versatility of a single hardware architecture. For instance, image processing tasks often require a large number of convolution operations and matrix computations, which are well-suited for execution on GPUs or TPUs. On the other hand, natural language processing tasks rely more on sequential data processing, which demands different memory bandwidth and processor architecture. Reinforcement learning tasks may require fast feedback mechanisms and low-latency computational environments. Therefore, a single hardware architecture is unlikely to meet the demands of all AI tasks, and the adaptability of hardware architectures becomes a major limiting factor for the expansion of AI applications. As AI application scenarios become more complex, hardware architectures must provide optimization across multiple dimensions, which a single architecture often cannot cover.

\subsubsection{Complexity of the System Software Stack and Lack of Optimization}
\textbf{Inadequate Development Toolchains.} Currently, AI accelerators and system software stacks (such as compilers, schedulers, libraries, etc.) do not fully exploit the potential of hardware, especially when dealing with emerging hardware architectures (e.g., RISC-V-based accelerators). For instance, RISC-V, despite its open-source nature offering great flexibility for customized hardware design, still lacks robust support in AI development toolchains when compared to traditional architectures like x86 or ARM. Many AI development tools, such as TensorFlow and PyTorch, offer support for existing GPUs and TPUs but provide limited support for emerging architectures like RISC-V. This results in developers facing steep learning curves and adaptation work when developing on these platforms. Additionally, current compilers and schedulers often fail to fully optimize hardware resources, preventing AI accelerators from reaching their full computational potential.

\textbf{Lack of Hardware-Software Synergy} There is a lack of tight synergy between existing hardware architectures and system software, particularly in emerging AI accelerators and system architectures (e.g., RISC-V-based designs), where hardware and software optimization is often disjointed, leading to suboptimal performance. The traditional separation between hardware and software development has caused a gap in the integrated development of hardware architectures and system software stacks, especially for new hardware platforms where hardware architecture design and software optimization tend to lag behind. For example, in the development of the TPU, Google's team conducted highly integrated optimization of both hardware architecture and software stack, enabling the TPU to maximize AI inference performance. However, for emerging AI accelerators based on open architectures like RISC-V, the deep integration of hardware and software design remains in the exploratory phase, preventing the level of cooperation seen in TPU development, which has contributed to performance bottlenecks.

\subsection{Opportunities and Example Solutions for Addressing These Issues}
By leveraging RISC-V-based AI accelerators, deep customization of Domain-Specific Architectures (DSA), and hardware-software co-optimization, the performance bottlenecks in AI hardware architectures can be effectively alleviated. The openness and customizability of RISC-V provide more efficient hardware solutions for AI tasks; DSA architectures address the limitations of traditional hardware by providing highly optimized hardware for specific tasks; and hardware-software co-optimization enhances computational efficiency through advanced compilers and schedulers. These technologies will help AI hardware architectures scale to meet the demands of large-scale training and inference tasks, improving computational power, reducing energy consumption, and offering greater flexibility and adaptability across diverse applications.

\subsubsection{RISC-V-Based AI Accelerators}

RISC-V, as an open-source Instruction Set Architecture (ISA), offers significant flexibility in AI hardware design, particularly for custom AI accelerators. It allows hardware architectures to be tailored to specific application scenarios, such as Convolutional Neural Networks (CNN), image processing, and Natural Language Processing (NLP), optimizing computational resources and power consumption. This customization addresses the issue of "insufficient heterogeneity and adaptability" in hardware architecture, enabling more efficient and energy-conserving processing units for specific tasks.

Unlike traditional ISAs (e.g., x86, ARM), RISC-V offers flexibility in adding custom instructions and hardware modules. Through the integration of accelerator cores (e.g., matrix multiplication units for deep learning) or specialized memory management units, RISC-V can significantly improve AI task performance, especially for image recognition, video processing, and NLP. Furthermore, RISC-V's openness eliminates vendor-locking, lowering development costs and accelerating technological innovation.

\textbf{Example Solutions:} SiFive's RISC-V-based AI accelerator provides customized hardware for CNNs and other machine learning tasks, offering higher computational density and lower latency compared to traditional GPUs, while maintaining similar power consumption. Alibaba's Xuantie RISC-V cores feature a customized instruction set to optimize AI inference, especially in edge computing, achieving 2-3 times the performance of CPUs and GPUs in image and video processing tasks. These RISC-V-based solutions effectively address performance bottlenecks in existing hardware (e.g., GPUs) when processing specific deep learning tasks, offering lower power consumption and higher computational density.

\subsubsection{Design and Optimization of Domain-Specific Architectures (DSA)}

Domain-Specific Architectures (DSA) provide hardware optimized for specific tasks such as AI inference or training. DSA architectures enhance performance and reduce power consumption by customizing processor cores, memory management, and data flow control for specific tasks. Unlike general-purpose hardware (e.g., CPUs or GPUs), DSA minimizes redundant computation, leading to significant performance improvements.

In AI, DSA architectures have shown great potential for both inference and training. Deep neural network (DNN) training requires extensive matrix multiplications, convolutions, and activation function computations. While GPUs excel at parallel computing, their general-purpose nature makes them less efficient for these specific tasks. DSA, by customizing hardware, reduces unnecessary computations, improving performance and energy efficiency.

\textbf{Example Solutions:} Google's Tensor Processing Unit (TPU) is a prime example of DSA, designed specifically for deep learning. TPUs feature custom matrix multiplication units and optimized memory bandwidth, achieving up to 3-5 times the training speedup compared to NVIDIA V100 GPUs with lower power consumption. NVIDIA's A100 GPU also uses specialized Tensor Cores to optimize matrix operations for deep learning. Intel's Nervana Neural Network Processor (NNP) employs DSA to enhance deep learning inference throughput, offering 5 times higher throughput and 2 times lower power consumption than traditional CPUs. These DSA-based solutions not only solve inefficiencies in traditional hardware but also enhance computational density, boosting AI model training and inference performance.

\subsubsection{Hardware-Software Co-Optimization in AI System Frameworks}

In AI systems, hardware-software co-design is essential for optimizing overall performance. Traditional architectures (e.g., GPUs) and system software (e.g., TensorFlow, PyTorch) often lack efficient co-optimization, leading to resource wastage and poor software scheduling. With the advent of customized hardware, such as RISC-V, achieving deep co-optimization between hardware and software has become a viable solution for improving hardware-software synergy.

AI frameworks like TensorFlow and PyTorch are increasingly optimized for specific hardware. For example, TensorFlow's XLA (Accelerated Linear Algebra) compiler generates machine code optimized for the hardware's characteristics (e.g., GPU, TPU, or RISC-V accelerators), improving hardware utilization. Additionally, TensorFlow is advancing towards low-precision computing (e.g., FP16, INT8), which reduces memory bandwidth requirements and enhances computational efficiency.

\textbf{Example Solutions:} TensorFlow's XLA compiler optimizes computation graphs based on the specific hardware (e.g., TPU, GPU, RISC-V), reducing unnecessary computational overhead. For example, when using TPUs, XLA optimizes convolution operations into matrix multiplications, significantly improving efficiency. In RISC-V-based AI accelerators, adding hardware acceleration units for specific instructions allows XLA to further optimize code, reducing latency and improving computational efficiency. Automated optimization tools, such as Intel's oneAPI and NVIDIA's CUDA Toolkit, also play a key role in improving co-optimization. These tools generate hardware-specific code based on the architecture, ensuring that RISC-V accelerators operate with optimal efficiency. This co-optimization allows RISC-V accelerators to outperform traditional CPUs and GPUs in specific AI tasks, such as convolution and matrix multiplication, with lower power consumption.

\subsection{Future Challenges to Tackle}
In the future, the main challenges in the field of AI hardware and software will include cross-architecture optimization and compatibility, balancing energy efficiency and cost for AI models, and achieving deep hardware-software co-optimization. Addressing these challenges will require innovations in hardware design to improve scalability and energy efficiency, as well as the development of more intelligent and flexible optimization strategies at the system software level. These efforts will provide solid technical support for large-scale AI model training and inference and promote the sustainable development of AI technologies.

\subsubsection{Cross-Architecture Optimization and Compatibility Issues}
\textbf{Challenges of Hardware Architecture Diversity.} With the parallel development of different hardware architectures such as RISC-V, ARM, and x86, the increasing demand for AI computations has led to significant hardware diversity. Although each architecture has its own advantages in specific tasks - such as RISC-V's customizability and openness, ARM's widespread use in mobile devices, and x86's generality and maturity - their differences present major challenges for cross-architecture optimization and compatibility.

The diversity of hardware architectures means that developers must create specialized code for each architecture, optimizing and adjusting it individually. This not only increases the workload for developers but also limits the potential for performance improvement. For example, deep learning frameworks like TensorFlow and PyTorch perform significantly differently across hardware platforms. While GPUs, TPUs, FPGAs, and RISC-V each have their strengths, interoperability among these platforms is often poor. Optimizing code for different hardware platforms requires significant time and effort, and the inefficiency in cross-platform migration results in wasted resources and reduced development efficiency.

As AI model sizes and complexity continue to grow, ensuring cross-architecture compatibility and optimization will be a major challenge for hardware and software co-evolution. This requires hardware designers to provide more unified interfaces at the architectural level, while software development must innovate to create cross-architecture programming languages and abstraction models, reducing redundant code and unnecessary performance losses between different hardware platforms.

\textbf{Lack of Unified Programming Models.} Currently, there is no unified programming model across AI hardware architectures, which means developers face substantial complexity and workload when migrating and optimizing code across multiple platforms. For example, the difference between the CUDA programming model for NVIDIA GPUs and TensorFlow optimization for TPUs requires developers to adjust code according to the characteristics of the hardware, leading to significant development time spent on migration and adaptation. Furthermore, with the rise of emerging architectures like RISC-V, the difficulty of cross-platform development is further exacerbated.

The key to addressing this challenge lies in developing a cross-architecture compatible unified programming model. Such a model would not only abstract away hardware differences but also provide developers with a simplified set of APIs and optimization strategies, thereby reducing the complexity of hardware adaptation. Intel's oneAPI is an example of an attempt to address this challenge by offering a unified programming model that supports development and optimization across various hardware platforms, from CPUs and GPUs to FPGAs. Although the range of supported hardware platforms is still limited, this approach offers a promising path for cross-architecture optimization.

\subsubsection{Balancing Energy Efficiency and Cost for Large-Scale AI Models}
\textbf{Scalability and Energy Efficiency of AI Hardware.} As AI models scale up (e.g., the widespread adoption of large pretrained models like GPT-3 and BERT), ensuring a balance between high performance, energy efficiency, and cost becomes a significant challenge for AI hardware. Training large-scale models requires vast computational resources, often involving thousands or even tens of thousands of GPUs working in parallel. However, as the model size increases, the demand for computational power leads to critical challenges in hardware scalability and energy efficiency.

For instance, training the GPT-3 model consumes approximately 1280 MWh of energy, equivalent to the electricity consumption of an average U.S. household for 10 years. This indicates that, despite the continuous improvement in AI hardware computing power, energy efficiency remains a significant issue in large-scale training. Existing hardware architectures (e.g., GPUs, TPUs) face the challenge of scaling computational power without significantly increasing power consumption. This requires hardware designs to focus not only on computational density but also on power management, thermal control, and resource scheduling.

To achieve a balance between energy efficiency and cost, future AI hardware designs will need to consider how to enhance scalability through innovative architectural approaches. For example, more efficient memory technologies (such as HBM2, DDR5) and refined power management strategies could lower energy consumption without compromising performance. Additionally, managing the computational load across different hardware components, reducing redundant calculations, and minimizing resource wastage will be critical challenges for future AI hardware design.

\textbf{Challenges in Energy Efficiency Optimization.} Despite significant progress in energy efficiency improvements in existing AI hardware, such as NVIDIA's A100 GPU, which has enhanced energy efficiency by using more advanced process nodes (7nm) and efficient Tensor Cores, the increasing scale of AI models and datasets presents new bottlenecks. For instance, as model sizes grow, memory bandwidth and latency become major limitations for energy efficiency, making it difficult to further scale computational power without significantly increasing power consumption.

In the future, energy efficiency optimization will face even greater challenges. As AI models continue to grow in size, how to improve energy efficiency through innovations in power management, more efficient algorithms (e.g., sparsity algorithms, low-precision computing), and optimization of hardware acceleration units will be crucial. Emerging technologies such as quantum computing and optical computing may offer new pathways for energy efficiency optimization in AI hardware, but their commercialization and practical applications will require further research and development.

\subsubsection{Deep Hardware-Software Co-Optimization}
\textbf{Deep Adaptation of AI Models to Hardware Architectures.} To fully exploit the potential of hardware, future AI hardware designs must pay closer attention to the specific needs of AI models, particularly in the context of large-scale model training and inference. Designing more efficient and specialized accelerators will be key to achieving high-performance computing. In this process, deep adaptation between AI models and hardware architectures will be crucial to enabling efficient computation.

Current hardware architectures, such as GPUs and TPUs, are primarily optimized for general deep learning tasks, but as AI applications diversify (e.g., image recognition, autonomous driving, NLP), a single hardware architecture can no longer meet the needs of all tasks. Future AI hardware designs will need to focus on developing specialized accelerators. For example, neural networks designed for image processing could benefit from more efficient convolutional accelerators, while NLP tasks may require hardware optimized for long short-term memory (LSTM) and Transformer architectures.

Particularly with RISC-V architecture, how to achieve deep integration between hardware and software will become a critical challenge. While RISC-V's customizability offers potential for this task, efficient adaptation to AI model requirements at the hardware level, as well as the extension of instruction sets and hardware modules to enhance computational efficiency, will require extensive research and innovation.

\textbf{Flexible Hardware Design and Co-Updated Software.} As new AI algorithms and tasks emerge, hardware must be more flexible to adapt to evolving demands. Simultaneously, system software stacks (e.g., compilers, schedulers, optimization tools) must continuously evolve to accommodate these new hardware architectures. For example, with the widespread use of low-precision computing (e.g., INT8, FP16) in AI inference, existing compilers and schedulers must automatically recognize and adapt to hardware features, generating optimized code.

Moreover, hardware design flexibility is essential for supporting different AI algorithms and tasks. Future hardware architectures will need to not only support existing deep learning frameworks but also accommodate new AI algorithms that may emerge, such as neuro-symbolic computing or meta-learning. This requires hardware architectures to possess dynamic configurability, enabling them to meet the demands of various computational tasks while ensuring seamless integration with evolving software stacks.

{\it Acknowledgments}. This work was supported by National Natural Science Foundation of China (NSFC) and Israel Science
Foundation (ISF) (Grant No. 62161146003 and 3698/21). Additional support was by National Natural Science Foundation of China (Grant No. 623B2006 and Grant No. 92464301).

%{\it Biography and Photo}. 

\bibliographystyle{JCST}
\bibliography{ref}

\label{last-page}
\end{multicols}
\label{last-page}
\end{document}